# Innovative Automated Stretch Elastic Waistband Sewing Machine for Garment Manufacturing

**Ray Wai Man Kong** [*1]

[*1]Adjunct Professor, City University of Hong Kong, Hong Kong

[*1]Modernization Director, Eagle Nice（International） Holding Ltd, Hong Kong

## ABSTRACT

There is applied research for the development of the Automated Stretch Elastic Waistband Sewing Machine represents a significant advancement in garment manufacturing, addressing the industry's need for increased efficiency, precision, and adaptability. This machine integrates innovative features such as a sensor-based automatic waistband expansion system, synchronized sewing speed and rolling wheel speed, and a differential feed top-loading mechanism. These enhancements streamline the sewing process, reduce manual intervention, and ensure consistent product quality. The machine's design incorporates both 3-wheel and 2-wheel rolling systems, each optimized for different elastic band dimensions and elongation factors. The 3-wheel rolling system accommodates a larger maximum boundary, while the 2-wheel rolling system offers a tighter operational range, providing flexibility to meet diverse manufacturing requirements. The Automated Stretch Elastic Waistband Sewing Machine has a design that controls the pulling apart force so as not to break the elastic waistband. It sets a new standard for quality and innovation, empowering manufacturers to meet the demands of a competitive market with precision and ease.

## I. INTRODUCTION

The garment manufacturing industry is undergoing a transformative shift driven by the increasing demand for efficiency, precision, and scalability. As global competition intensifies, manufacturers are compelled to adopt innovative solutions that enhance productivity while maintaining high-quality standards. One such advancement is the development of the Innovative Automated Stretch Elastic Waistband Sewing Machine, a technology poised to revolutionize the garment production process.

Traditional sewing methods for elastic waistbands are labour-intensive and prone to inconsistencies, often resulting in variable product quality and increased production times. These challenges are exacerbated by the growing consumer demand for stretchable and comfortable garments, which require precise and consistent sewing techniques. The introduction of automation in this domain addresses these issues by streamlining the sewing process, reducing manual intervention, and ensuring uniformity across products.

The Innovative Automated Stretch Elastic Waistband Sewing Machine leverages cutting-edge technology to automate the intricate task of sewing elastic waistbands, offering several key benefits. It enhances production speed, allowing manufacturers to meet tight deadlines and scale operations efficiently. The machine's precision reduces material waste and minimizes errors, leading to cost savings and improved profit margins. Additionally, automation alleviates the reliance on skilled labour, addressing labour shortages and enabling a more flexible workforce deployment.

By integrating this advanced technology, garment manufacturers can achieve a competitive edge in the market, responding swiftly to fashion trends and consumer preferences. The automation of elastic waistband sewing not only optimizes production processes but also sets a new standard for quality and innovation in the industry. As the garment sector continues to evolve, embracing such technological advancements will be crucial for manufacturers to aim to thrive in a rapidly changing landscape.

## II. FUNCTION OF AUTOMATED STRETCH ELASTIC WAISTBAND SEWING MACHINE

The Automated Stretch Elastic Waistband Sewing Machine is equipped with a range of advanced features designed to enhance efficiency, precision, and adaptability in garment manufacturing. Here are some of its key functions:

(a). Automatic Waistband Expansion:
The machine incorporates a sensor-based automatic waistband expansion system, eliminating the need for traditional knee-control switches. This innovation provides a faster and more convenient operation, streamlining the sewing process and reducing operator fatigue.

(b). Differential Feed Top-Loading:
The differential feed top-loading system ensures smooth and swift fabric feeding, effectively preventing fabric layering issues during the attachment of upper and lower seams. This function is crucial for maintaining fabric integrity and achieving high-quality finishes.

(c). Three-Wheel Positioning and Circular with alignment wheel Operation:
The innovative three-wheel positioning and circular operation mechanism enable the machine to stitch fabric smoothly, ensuring consistent edge alignment and even tension. The alignment wheel can align the sewing line. This results in visually appealing stitches and reduces the likelihood of errors.

(d). Precision Elastic Attachment:
The machine addresses the challenges of achieving neatly aligned and evenly tensioned elastic attachments by replacing traditional manual stitching with automated precision. This ensures consistent quality and reduces the need for work.

(e). Advanced Fabric Alignment Technology:
Featuring advanced patent technology for automated fabric alignment, the machine ensures smooth and seamless operation. This technology enhances the machine's ability to handle various fabric types, making it highly adaptable to diverse manufacturing needs.

(f). Enhanced Adaptability:
Compared to similar products, this elastic waistband sewing machine offers higher adaptability, catering to a wide range of fabric types and manufacturing requirements. This versatility makes it an ideal choice for manufacturers looking to expand their product range.

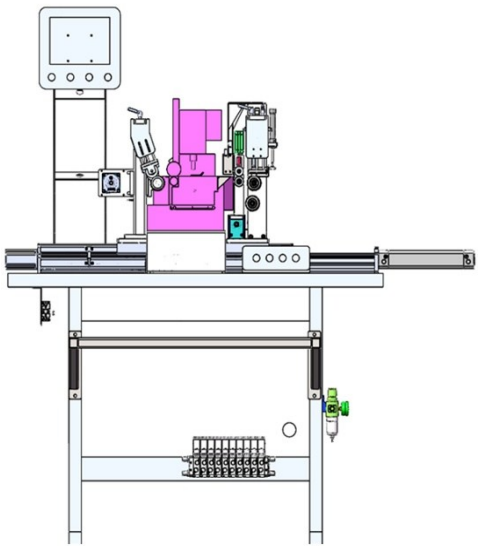
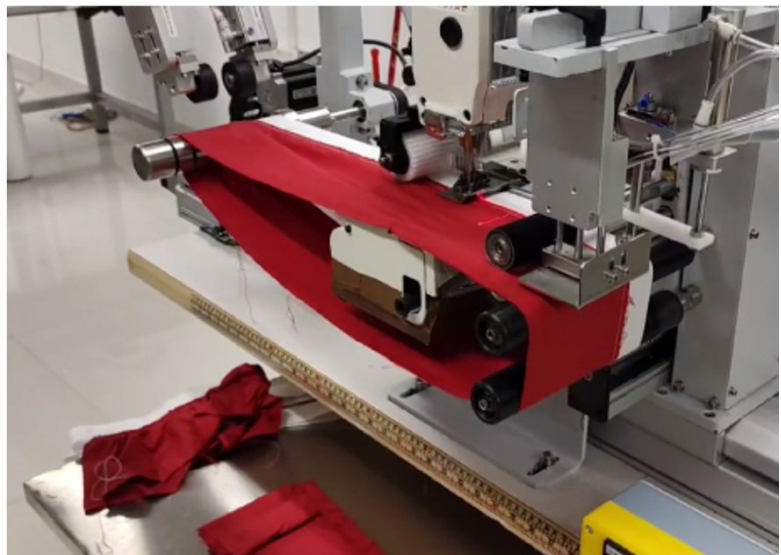

**Figure 1** Outlook of Automated Stretch Elastic Waistband Sewing Machine

By integrating these advanced functions, the Automated Stretch Elastic Waistband Sewing Machine sets a new standard for efficiency and quality in garment manufacturing, enabling producers to meet the demands of a competitive market with precision and ease as shown in the automated machinery design in Fig. 1.

The output of elastic band sewing is the waistband for the woven garment. Introduction to Garment Waistbands for Dresses and Pants

Garment waistbands are essential components in the design and functionality of dresses and pants, serving both aesthetic and practical purposes. They play a crucial role in ensuring a comfortable fit, enhancing the overall silhouette, and providing structural integrity to the garment. This introduction explores the various types of waistbands used in dresses and pants, their construction techniques, and their significance in modern fashion.

Types of Waistbands
Elastic Waistbands:

- Description:
  Elastic waistbands are made from stretchable materials that provide flexibility and comfort. They are commonly used in casual wear, activewear, and children's clothing.
- Application:
  These waistbands allow for easy dressing and undressing, making them ideal for garments that require a relaxed fit, such as joggers, leggings, and casual dresses.

Flat Waistbands:
- Description:
  Flat waistbands are typically made from the same fabric as the garment and are sewn into the waistline. They can be finished with a facing or lining for added comfort and style.
- Application:
  Commonly found in tailored pants, skirts, and dresses, flat waistbands provide a polished look and can be designed with various closures, such as buttons, zippers, or hooks.

Contoured Waistbands:
- Description:
  Contoured waistbands are shaped to follow the natural curves of the body, providing a more tailored fit. They are often used in fitted garments to enhance the silhouette.
- Application:
  These waistbands are popular in dresses and high-waisted pants, offering both comfort and style while preventing gaping at the waist.

Drawstring Waistbands:
- Description:
  Drawstring waistbands feature a cord or ribbon that can be adjusted to achieve the desired fit. They are often used in casual and sporty designs.
- Application:
  Common in shorts, sweatpants, and some dresses, drawstring waistbands provide versatility and comfort, allowing wearers to customize their fit.

Belted Waistbands:
- Description:
  Belted waistbands incorporate a belt or sash that can be tied or fastened to cinch the waist. This style adds a decorative element while enhancing the garment's fit.
- Application:
  Frequently seen in dresses and high-waisted pants, belted waistbands can create a flattering hourglass silhouette and add visual interest to the design.

Construction Techniques
The construction of waistbands can vary based on the type and design of the garment. Key techniques include:

Sewing Methods:
Traditional sewing methods involve manual stitching, which can be labour-intensive and may lead to inconsistencies. However, advancements in technology, such as the Innovative Automated Stretch Elastic Waistband Sewing Machine, are revolutionizing this process by automating the sewing of elastic waistbands, ensuring precision and uniformity.

Finishing Techniques:
Waistbands can be finished with various techniques, including topstitching, binding, or adding interfacing for added stability. These finishing touches enhance the durability and appearance of the waistband.

Sizing and Adjustments:

Proper sizing is crucial for waistbands to ensure a comfortable fit. Manufacturers often incorporate sizing charts and adjust patterns to accommodate different body shapes and sizes.

Significance in Modern Fashion:
Waistbands are not just functional elements; they are integral to the overall design and style of dresses and pants. In modern fashion, the choice of waistband can influence the garment's silhouette, comfort, and versatility. As consumer preferences shift towards comfort and flexibility, the demand for innovative waistband solutions, such as elastic and adjustable designs, continues to grow.

Moreover, the integration of advanced technologies in waistband production, such as automation, allows manufacturers to respond quickly to changing fashion trends while maintaining high-quality standards. This adaptability is essential in a competitive market where speed and efficiency are paramount.

In summary, garment waistbands for dresses and pants in Fig. 2 are vital components that enhance both the functionality and aesthetic appeal of clothing. With various types and construction techniques available, waistbands play a significant role in achieving the desired fit and style. As the garment industry evolves, embracing innovative solutions for waistband production will be crucial for manufacturers aiming to meet consumer demands and stay competitive in the ever-changing fashion landscape.

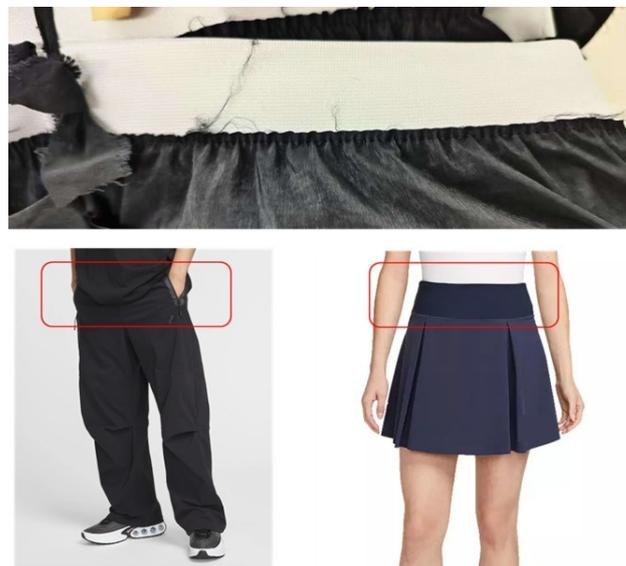

**Figure 2** Elastic Waistband Sewing Part and Garment Diagram

## III. Literature Review

A literature review from Groover, M. P. [1] on the development of automatic machinery, particularly in the context of manufacturing, reveals a rich history of innovation aimed at enhancing efficiency, precision, and adaptability.

1. Historical Evolution of Automation:
   Early developments in automation focused on mechanizing simple, repetitive tasks to increase production speed and reduce labour costs. The Industrial Revolution marked a significant leap with the introduction of steam-powered machinery, which laid the groundwork for modern automation.
2. Advancements in Control Systems:
   The integration of electronic control systems in the mid-20th century revolutionized automation. Programmable logic controllers (PLCs) and computer numerical control (CNC) systems enabled more complex and precise operations, allowing for greater flexibility and customization in manufacturing processes.
3. Robotics and Intelligent Systems:
   Siciliano, B., & Khatib, O. (Eds.) [2] mentioned the advent of robotics has been a game-changer in automation, with robots taking on tasks ranging from assembly to quality inspection. Recent advancements in artificial intelligence (AI) and machine learning have further enhanced robotic capabilities, enabling adaptive and intelligent decision-making in real time.
4. Sensor Technology and IoT:

Lee, J., Bagheri, B., & Kao, H. A. [3] uses the development of advanced sensor technologies and the Internet of Things (IoT) has facilitated the creation of smart manufacturing environments. Sensors provide critical data for monitoring and optimizing machine performance, while IoT connectivity allows for seamless communication between devices and systems.
5. Automation in Textile and Garment Manufacturing:
In the apparel and garment industry, Ray automation has focused on processes such as cutting, sewing, and finishing. Innovations like automated sewing machines and fabric handling systems have significantly reduced manual labour and improved product consistency.
6. Challenges and Opportunities:
Despite the benefits, the adoption of automation in manufacturing faces challenges such as high initial costs, the need for skilled personnel to manage and maintain automated systems, and concerns about job displacement. However, the potential for increased productivity, quality, and flexibility presents significant opportunities for growth and innovation.
7. Future Trends:
Based on the Lean Methodology For Garment Modernization from Prof Dr Ray WM Kong [4], the future of automatic machinery development is likely to be shaped by advancements in AI, machine learning, and robotics.

The machinery to implementation can refer to the Mixed-Integer Linear Programming (MILP) for Garment Line Balancing for the optimized resource and the Design and Experimental Study of Vacuum Suction Grabbing Technology to Grasp Fabric Pieces can support the loading of fabric pieces to the machinery with Innovative Vacuum Suction-grabbing Technology for Garment Automation. In K. M. Batoo (Ed.), Science and Technology: Developments and Applications from Prof Dr Ray WM Kong [5] [6] [7].

The focus will be on creating more autonomous systems capable of self-optimization and adaptation to changing production demands.

## IV. TECHNICAL POINTS IN THE DESIGN OF THE AUTOMATED STRETCH ELASTIC WAISTBAND SEWING MACHINE

The technical points in the design of the automated stretch elastic waistband sewing machine include the following:

**Automated Machinery Structure**
In Fig. 3, the Semi-enclosed lead screw slide rail is used to support the stretching mechanism and the deviation correction mechanism. Guarantee the stability of the stretch.
On the left, a servo motor is used to stretch the scarf, and the servo motor can actively stop when the fabric piece is overloaded, to protect the fabric piece. It will not be damaged by excessive tensile force.

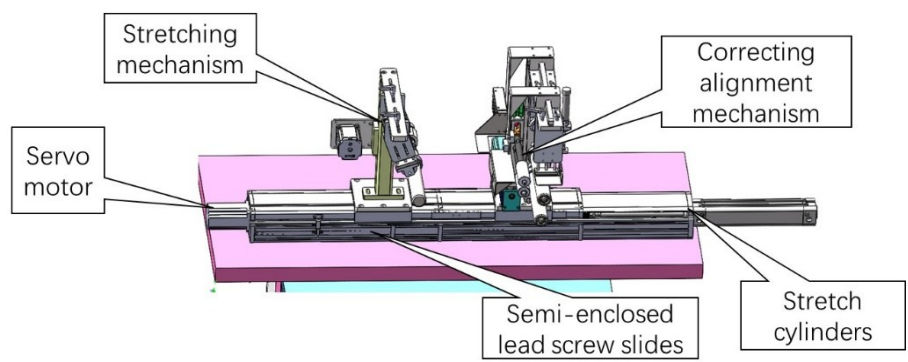

**Figure 3** Automated Stretch Elastic Waistband Sewing Machine Diagram

On the right, the stretch cylinder is used to pull the correction mechanism, and when the needle and thread are replaced in the sewing machine, the correcting alignment mechanism can be pulled out, so that the thread can be better changed and maintained.

**Stretching mechanism**
The stretching mechanism in Fig. 4 uses the servo motor to drive the pulling force of the moving right and left wheels. The elastic band holds to the moving wheels. The moving right wheel moves in the right direction and the moving left wheel moves in the left direction. The opposite direction forms the pulling force to pull apart of elastic band. The elastic band is an elastic characteristic which can elongate to extend the length when the pull-apart force is good enough to act on the elastic band.
During the experiment from Company A, the elongation factor is not a constraint, which has been reduced from 2.72 to 2.25 when pulling out the elasticity of the elastic band to measure the elongation as shown in Fig 4.

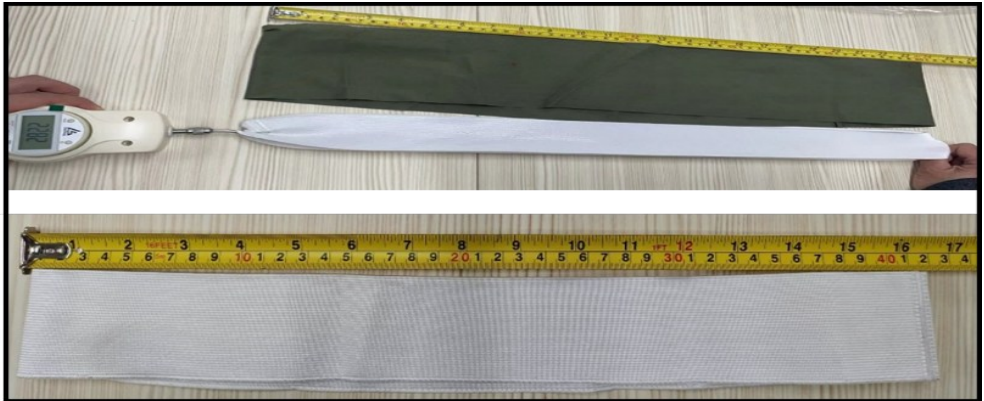

**Figure 4** Elastic Band Pulling-apart Test to measure the elongation

The test experiment is useful in the design of an automated machine based on the elastic band elongation and work requirements. Measurements of elongation are important in a variety of industries and applications, such as mechanical engineering, medicine, and construction. The formula for elongation is shown in the following:

$$\varepsilon = (\Delta L/L) \times 100 \tag{1}$$

Where is the elongation percentage, ΔL is the change in length, and $L$ is the original length.

Elongation ε = (Final Length - Original Length) / Original Length x 100%
The maximum acceptable elongation before breaking of the elastic band. The elastic band has been sewn in a round shape. The elongation measurement refers to the case study and experiment.

$\varepsilon_{max}$ = (610mm – 420mm)/420 mm x 100%
$\varepsilon_{max}$ = 45.2%

Applied Force by Hooke's Law, $F_s = ks$, where k is a factor characteristic of the elastic band (i.e., its stiffness), and s is small compared to the total possible deformation of the elastic band.
$F_s$ = k $\Delta s$
$F_s$ = 22.82N
k= $F_s$ / $\Delta s$, $\Delta s$ =(610mm - 420mm)
k = 22.82N / 190mm
k = 120N/m  (for constraint factor of the elastic band)
Different types of elastics have various deformations.

The *k* is not a constant factor characteristic of the elastic band in the case and experiment, the experiments in the 10.1 Force-Deformation Curve from the Pressbook of British Columbia/Yukon Open Authoring Platform [1] have shown that the change in length ($\Delta L$) depends on only a few variables. As already noted, $\Delta L$ is proportional to force *F* and depends on the substance from which the material is made. Additionally, the change in length is proportional to the original length $L_0$ and inversely proportional to the cross-sectional area of the elastic band. For example, a long guitar string will stretch more than a short one, and a thick string will stretch less than a thin one. We can combine all these factors into one equation for ΔL:

$$\Delta L = \frac{1}{Y} \frac{F}{A} L_0 \tag{2}$$

Where $\Delta L$ is the change in length, *F* is the applied force, *Y* is a factor, called the elastic modulus or Young's modulus, that depends on the substance, *A* is the cross-sectional area, and $L_0$ is the original length.

In Fig.5, the straight segment is the linear region where Hooke's law is obeyed. The slope of the straight region is 1 / k. For larger forces, the graph is curved but the deformation is still elastic—ΔL will return to zero if the force is removed.

It has still greater forces that permanently deform the object until it finally fractures. The shape of the curve near fracture depends on several factors, including how force F is applied. In Fig. 5, the slope increases just before fracture, indicating that a small increase in F is producing a large increase in L near the fracture of elastic bands or other objects.

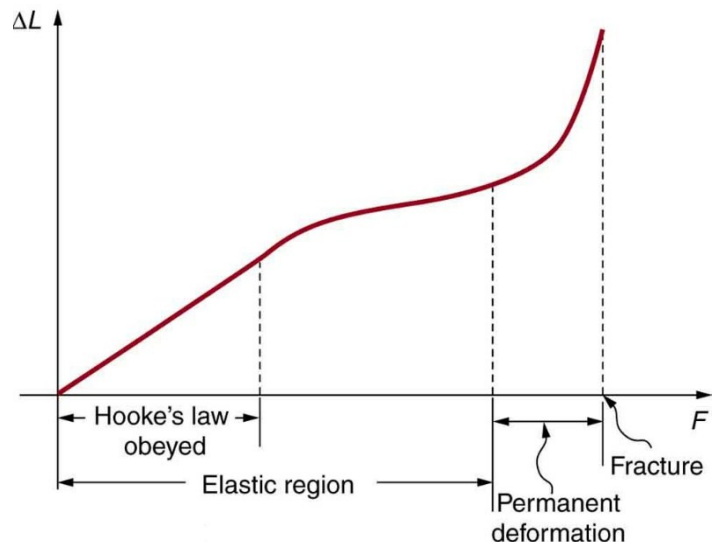

**Figure 5** A graph of deformation ΔL versus applied force F

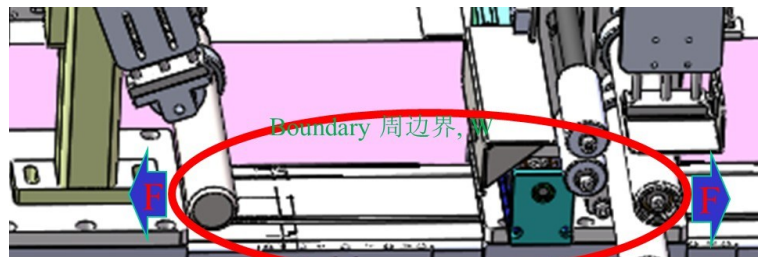

**Figure 6** Pulling-apart force analysis in the machinery diagram

In Fig. 6, the pulling-apart force analysis in the machinery diagram calculates the maximum acceptable pulling-apart force, not to break the elastic band for the design of automatic machine specification.

## Automated Machinery Design

The innovative automated stretch elastic waistband sewing machine design for garment manufacturing considers the practical waistband size. The design of automatic machines should satisfy the waistband requirements. The machine has been designed the 3 wheels and 2 wheels rotating and pulling apart mechanism to satisfy all sizes of the waist for the styles for women and men.

The capability of machinery is required to satisfy the minimum size and maximum size of the waist. Machineability is the critical thing which defines the fabrication of machines in the machining process of cutting, shaping, or removing material from a workpiece using a machine tool. All our machining is done with the highest quality machine tools. The electricity circuit design is supported to control the automated machine for the automation.

## Automated Stretch Elastic Waistband Sewing Machine Design

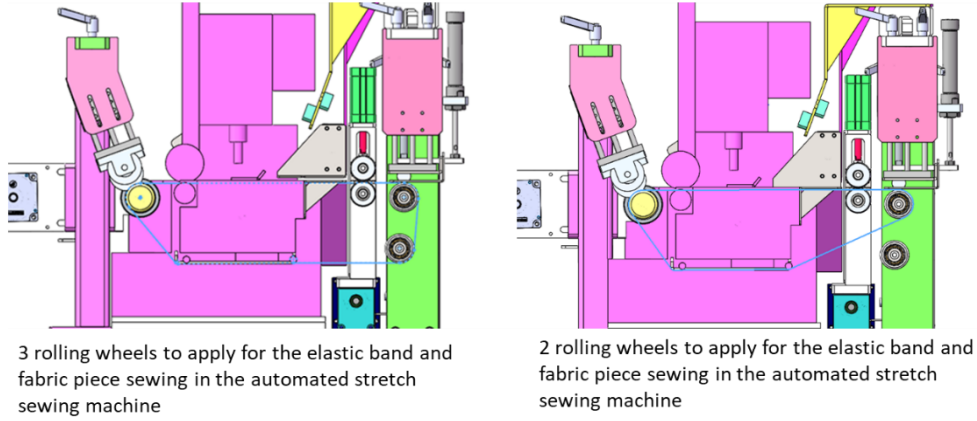

**Figure 7** Design of 3 wheels and 2 wheels of machinery diagram

The combination of design of both 2 wheels and 3 wheels in the waistband sewing machine can satisfy all waist size requirements in garment manufacturing successfully.

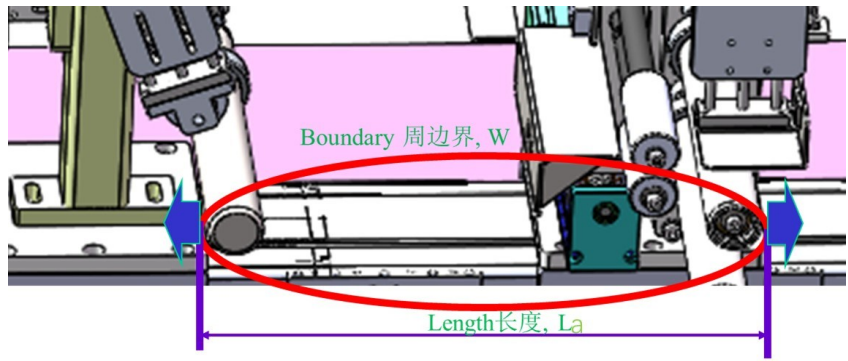

**Figure 8** Relationship diagram of Rolling wheel and waistband diagram

### Automated Machinery Design for 3 Rolling Wheels

The length of $L_a$ between the right rolling wheel and left rolling wheel in Fig. 7 and Fig. 8 defines the $L_x$ length between two rolling wheels. The round shape of the elastic band's dimension defines $W_x$ in Fig. 7. The maximum of $W_x$ is driven by the maximum $L_x$. The relationship is the maximum elastic band $W_x$ and $L_x$ is shown in the following:

$$Max_{3wheels}(W_x) = Max_{3wheels}(L_x)\, E_{xa} \qquad (3)$$

where the $Max_{3wheels}(W_x)$ is the maximum of the acceptable rounded boundary of the elastic band, $Max_{3wheels}(L_x)$ is the maximum of acceptable length between wheels. $E_{xa}$ defines the elongation factor.

$Max_{3wheels}(L_x)$ = 750mm
$Max_{3wheels}(W_x) = 1691mm$
$E_{xa}$ = 2.25

$$Min_{3wheels}(W_x) = Min_{3wheels}(L_x)\, E_{xb} \qquad (4)$$

where the $Min_{3wheels}(W_x)$ is the minimum of the acceptable rounded boundary of the elastic band, $Min_{3wheels}(L_x)$ is the minimum of acceptable length between wheels. $E_{xb}$ defines the elongation factor.

$Min_{3wheels}(L_x)$ = 300mm
$Min_{3wheels}(W_x) = 861mm$
$E_{xb}$ = 2.72

The elongation factor is not a constraint, which has been reduced from $E_{xb}$, 2.72 to $E_{xa}$, 2.25 when pulling apart the elasticity of the elastic band for the 3 rolling wheels in the automated machinery design.

## Automated Machinery Design for 2 Rolling Wheels

The length of $L_a$ between the right rolling wheel and left rolling wheel in Fig. 7 and Fig. 8 defines the $L_x$ length between two rolling wheels. The round shape of the elastic band's dimension defines $W_x$ in Fig. 7. The maximum of $W_x$ is driven by the maximum $L_x$. The relationship is the maximum elastic band $W_x$ and $L_x$ is shown in the following:

$$Max_{2wheels}(W_x) = Max_{2wheels}(L_x)\, E_{xc} \qquad (5)$$

where the $Max_{2wheels}(W_x)$ is the maximum of the acceptable rounded boundary of the elastic band, $Max_{2wheels}(L_x)$ is the maximum of acceptable length between wheels. $E_{xc}$ defines the elongation factor.

$Max_{2wheels}(L_x)$ = 750mm
$Max_{2wheels}(W_x) = 1619mm$
$E_{xc}$ = 2.15

$$Min_{2wheels}(W_x) = Min_{2wheesl}(L_x)\, E_{xd} \qquad (6)$$

where the $Min_{2wheels}(W_x)$ is the minimum of the acceptable rounded boundary of the elastic band, $Min_{2wheels}(L_x)$ is the minimum of acceptable length between wheels. $E_{xd}$ defines the elongation factor.

$Min_{2wheels}(L_x)$ = 300mm
$Min_{2wheels}(W_x) = 750mm$
$E_{xd}$ = 2.50

The elongation factor is not a constraint, which has been reduced from $E_{xd}$, 2.50 to $E_{xc}$, 2.15 when pulling apart the elasticity of the elastic band for the 3 rolling wheels in the automated machinery design.

## V. METHODOLOGY IN DESIGN OF AUTOMATED STRETCH ELASTIC WAISTBAND SEWING MACHINE

The main methodology is to study the practical requirements for production. Referring to Prof Dr Ray Wai Man Kong's article, Lean Methodology for Garment Modernization, industrial engineering and lean study are required to study the whole garment manufacturing process flow. Industrial engineering studies in requirement study play a crucial role in building automation of garment manufacturing. An industrial engineer is working for the Here's how it is utilized:

(a)     Work Measurement: Industrial engineering study involves conducting time and motion studies to measure the time taken to perform each operation in the sewing assembly line. This data is essential for calculating each operation's cycle time and determining each operator's capacity.

(b)     Standardized Work Methods: Industrial engineers analyze the work methods used in garment manufacturing and identify opportunities for improvement. They develop standardized work methods that optimize efficiency and reduce variability in the production process. These standardized methods contribute to effective automation.

(c)     Garment Process Analysis: Industrial engineers analyze the entire garment manufacturing process, from receiving raw materials to the final product's shipment. The garment sewing process bottlenecks, inefficiencies, and areas of improvement can be identified by Value Stream Mapping. By understanding the process flow, the future state of VSM can identify opportunities for automation and optimize the sequence of operations.

(d)     Capacity and Manpower Resource Allocation: Industrial engineers assess the garment sewing workforce and equipment available in the garment manufacturing facility. They determine the number of operators required for each operation based on the calculated cycle time and operator capacity. This helps in allocating resources effectively and achieving a balanced line.

(e)      Layout Design: Industrial engineers consider the layout of the sewing assembly line and its impact on efficiency. They analyze the flow of materials, equipment placement, and operator movement. By optimizing the layout including (1) batch layout, (2) one-piece line layout, (3) conveyor line layout and (4) intelligent line layout, besides the batch layout, other 3 production line layouts for the sewing process can minimize unnecessary movement, reduce transportation time, and improve overall line balancing and automation.

(f) Ergonomics and Workplace Design: Industrial engineers consider ergonomics principles to design workstations that promote worker safety, comfort, and productivity. They ensure that the layout and design of workstations support efficient movement and minimize fatigue. This contributes to improved automation by enhancing operator performance.

(g) Continuous Improvement: Industrial engineering study emphasizes continuous improvement in automation as referred to the Lean Methodology for Garment Modernization. Industrial engineers monitor the line's performance, collect data, and analyze it to identify areas for further optimization. They implement changes, conduct follow-up studies, and refine the line-balancing process to achieve higher efficiency and productivity.

By utilizing industrial engineering study in the automation of garment manufacturing, companies can optimize their production processes, reduce lead times, improve resource utilization, and enhance overall efficiency. This results in increased productivity, cost savings, and improved customer satisfaction as referred to in the methodology of Lean Methodology in Modern Garment Manufacturing and Line Balancing in the Modern Garment Industry from Prof Dr Ray WM Kong.

*Applied the Automated Machinery Design for both 3 Rolling Wheels and 2 Rolling Wheels*

The new automated stretch elastic waistband sewing machine has applied the 3 Rolling Wheels and the 2 Rolling Wheels in the automated machinery design. The flexibility for the application is greater than either the 3 Rolling Wheels design or the 2 Rolling Wheels design. After the combination design of rolling wheels in the singular machine, the maximum $W_x$ of 3 rolling wheels design and the minimum $W_x$ of 2 rolling wheels design.

$$Max_w(W_x) = Max_{3wheels}(L_x)\, E_{xa} \qquad (7)$$

$$Min_w(W_x) = Min_{2wheels}(L_x)\, E_{xd} \qquad (8)$$

where the $Max_w(W_x)$ is the maximum of the acceptable rounded boundary of the elastic band, $Min_w(L_x)$ is the minimum of acceptable length between wheels. There is a great solution to maximize and to minimize the elastic band for sewing garments.

## VI. Force Analysis For The Automated Stretch Elastic Waistband Sewing Machine

The force analysis of the prevention of the break of the elastic band has studied the formula of force analysis by over-pulling force from the machine; the machine has set the electricity force sensor and controller to control the stress force, not over the break force of the elastic band for waistband sewing operation as shown in Fig. 9.

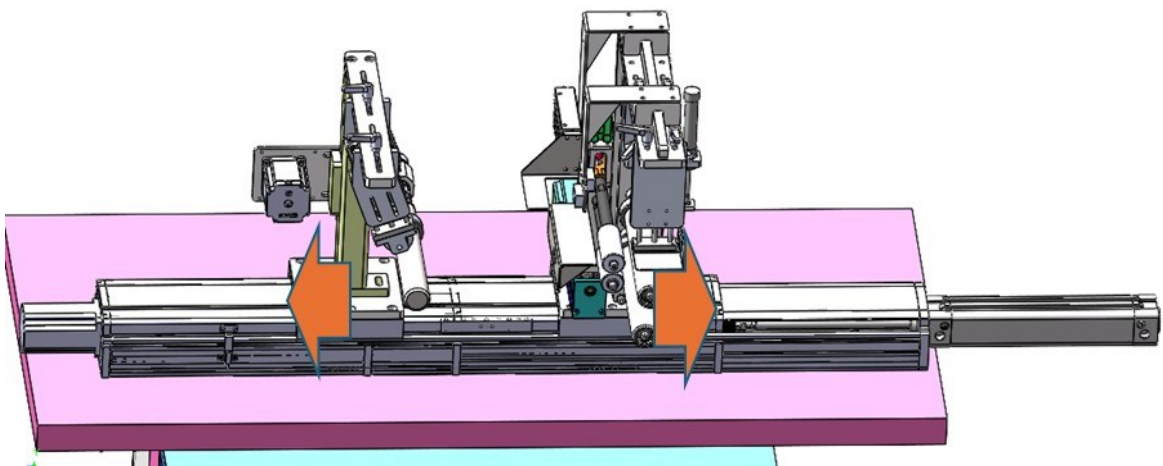

**Figure 9** Pulling apart force analysis to maximize a waistband diagram.

The applied force is controlled not to break the elastic waistband by the automated stretch elastic waistband sewing machine. The applied force is to stretch the elastic waistband for 610mm half round of the waist. The break force of the elastic band is 31N based on the experiment.

   Applied Force = 22.82 N
   Safety Force ≤ 31N (assume break of an elastic band*)
   Torque = Force x Length $T = F * \vec{r}$

$$Force, F = \frac{T}{\vec{r}} \tag{9}$$

Torque from the machine's motor, $Torque\ T_s = 2.4Nm$,
Controllable torque is required to limit the machinery torque not over the break of elastic waistband.

$$Limited\ Torque\ T_s = T\ (C_\%) \tag{10}$$

Where $T_s$ is the limited torque, $C_\%$ is a setup control % in the machinery controller, not allowing over force to break down the elastic band in the machine,

$$Limited\ Force\ F_s = \frac{T\ (C_\%)}{r} \tag{11}$$

where $F_s$ is the limited force not to break the waistband,

Limited $Force\ F_s$=2.4Nm $(C_\%)$ / r
Limited $Force\ F_s$ ≤ 31N (based on waistband strength in the experiment of case study)
where r is the radius of the servo motor's rod,
 r = 9.5mm
It calculated the control percentage of the motor, $C_\%$.
Hence, $C_\% ≤ 12\%$ ,
It set the $C_\% = 12\%$ the Limited $Forec\ F_s = 30.3N$.

In Fig 10, the Elasticity control schematic diagram shows the controller how to control the limited length, $L_x$ and the limited torque and limited force not to break the elastic waistband.

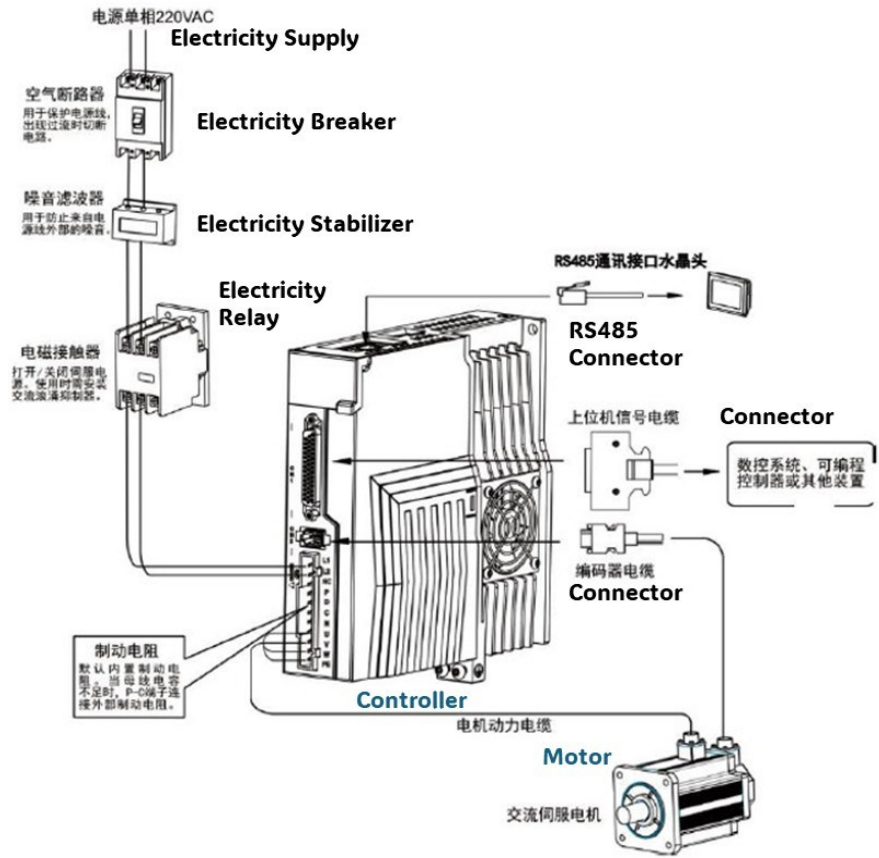

**Figure 10** Elasticity control schematic diagram

## VII. Conclusion

The analysis of automated machinery design for both 3 rolling wheels and 2 rolling wheels provides insights into the relationship between the length between wheels ($L_x$), the maximum and minimum acceptable rounded boundary of the elastic band ($W_x$), and the elongation factors ($E_{xa}, E_{xb}, E_{xc}, E_{xd}$). Here are the key conclusions:

a) Elongation Factor Reduction:
   For both the 3-wheel and 2-wheel designs, the elongation factor is reduced when the elasticity of the elastic band is pulled apart. Specifically, the elongation factor decreases from $E_{xb}$ (2.72) to $E_{xa}$ (2.25) for the 3-wheel design and from $E_{xd}$ (2.50) to $E_{xc}$ (2.15) for the 2-wheel design. This reduction indicates that the system is designed to accommodate a certain level of elasticity without compromising the structural integrity or performance of the machinery.

b) Maximum and Minimum Acceptable Boundaries:
   The maximum and minimum acceptable rounded boundaries ($W_x$) for the elastic band are determined by the maximum and minimum lengths between wheels ($L_x$) and their respective elongation factors. For the 3-wheel design, the maximum $Max_{3wheels}(W_x)$ is 1691mm, and the minimum $Min_{3wheels}(W_x)$ is 861mm. For the 2-wheel design, the maximum $Max_{2wheels}(W_x)$ is 1619mm, and the minimum $Max_{2wheels}(L_x)$ is 750mm. These values reflect the design constraints and capabilities of each system.

c) Comparison Between 3-Wheel and 2-Wheel Designs:
   The 3-wheel design allows for a slightly larger maximum acceptable boundary ($W_x$) compared to the 2-wheel design, suggesting that it can handle larger elastic bands or greater elasticity. However, the 2-wheel design has a slightly higher minimum acceptable boundary, indicating a tighter range of operation.

d) Design Flexibility:
   The ability to adjust the elongation factor without it being a constraint suggests that the machinery is designed with flexibility in mind, allowing it to adapt to different elastic band properties and manufacturing requirements.

Overall, the design considerations for both the 3-wheel and 2-wheel systems highlight the importance of balancing elasticity, wheel spacing, and elongation factors to achieve optimal performance in automated machinery for elastic band processing in the design of Automated Stretch Elastic Waistband Sewing Machines for Garment Manufacturing as well as refer the Design a New Pulling Gear for the Automated Pant Bottom Hem Sewing Machine, International Research Journal of Modernization in Engineering Technology and Science and cross-industrial analysis to study the detail of operation process and optimization from Prof Dr Ray WM Kong [8] [9].

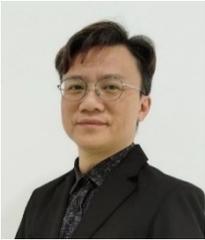

**Ray Wai Man Kong** (Senior Member of IEEE, member of IET, member of Public Administration Association) Hong Kong, China. He received a Bachelor of General Study degree from the Open University of Hong Kong, Hong Kong in 1995. He received an MSc degree in Automation Systems and Engineering and an Engineering Doctorate from the City University of Hong Kong, Hong Kong in 1998 and 2008 respectively.

From 2005 to 2013, he was the operations director with Automated Manufacturing Limited, Hong Kong. From 2020 to 2021, he was the Chief Operating Officer (COO) of Wah Ming Optical Manufactory Ltd, Hong Kong. He is a modernization director with Eagle Nice (International) Holdings Limited, Hong Kong. He is an Adjunct Professor of the System Engineering Department at the City University of Hong Kong, Hong Kong. He published more articles in international journals on robotic technology, grippers, automation and intelligent manufacturing. His research interests focus on intelligent manufacturing, automation, maglev technology, robotics, mechanical engineering, electronics, and system engineering for industrial factories.

Prof. Dr. Kong Wai Man, Ray is Vice President of CityU Engineering Doctorate Society, Hong Kong and a chairman of the Intelligent Manufacturing Committee of the Doctors Think Tank Academy, Hong Kong. He has published more intellectual properties and patents in China.